\RequirePackage{xcolor}
\documentclass{article}

\usepackage{microtype}
\usepackage{graphicx}
\usepackage{subfigure}
\usepackage{booktabs} % for professional tables
    
\usepackage{hyperref}
\usepackage{amsfonts}
\usepackage{listings}
\usepackage{xcolor}

\definecolor{codegreen}{rgb}{0,0.6,0}
\definecolor{codegray}{rgb}{0.5,0.5,0.5}
\definecolor{codepurple}{rgb}{0.58,0,0.82}
\definecolor{backcolour}{rgb}{0.95,0.95,0.92}

\lstdefinestyle{pythonstyle}{
frame = single,
commentstyle=\color{codegreen},
keywordstyle=\color{magenta},
numberstyle=\tiny\color{codegray},
stringstyle=\color{codepurple},
basicstyle=\ttfamily\footnotesize,
breakatwhitespace=false,         
breaklines=true,                 
captionpos=b,                    
keepspaces=true,                 
numbers=left,                    
numbersep=5pt,                  
showspaces=false,                
showstringspaces=false,
showtabs=false,                  
tabsize=2,
}
\usepackage[accepted]{icml2020}
\usepackage{amsmath}
\usepackage{makecell}
\usepackage{tcolorbox}
\newcommand{\highlight}[1]{\colorbox{blue!10}{#1}}

\icmltitlerunning{PointMask}

\begin{document}

\twocolumn[
\icmltitle{PointMask: Towards Interpretable and Bias-Resilient Point Cloud Processing}



\icmlsetsymbol{equal}{*}

\begin{icmlauthorlist}
\icmlauthor{Saeid Asgari Taghanaki}{atoai}
\icmlauthor{Kaveh Hassani}{equal,atoai}
\icmlauthor{Pradeep Kumar Jayaraman}{equal,ator}
\icmlauthor{Amir Hosein Khasahmadi}{atoai}
\icmlauthor{Tonya Custis}{asaai}
\end{icmlauthorlist}

\icmlaffiliation{atoai}{Autodesk AI Lab, Toronto, Canada}
\icmlaffiliation{ator}{Autodesk Research, Toronto, Canada}
\icmlaffiliation{asaai}{Autodesk AI Lab, San Francisco, USA}

\icmlcorrespondingauthor{Saeid Asgari}{saeid.asgari.taghanaki@autodesk.com}
% \icmlcorrespondingauthor{Eee Pppp}{ep@eden.co.uk}

\icmlkeywords{Machine Learning, ICML}

\vskip 0.3in
]



\printAffiliationsAndNotice{\icmlEqualContribution} % otherwise use the standard text.

\begin{abstract}
Deep classifiers tend to associate a few discriminative input variables with their objective function, which in turn, may hurt their generalization capabilities. To address this, one can design systematic experiments and/or inspect the models via interpretability methods. In this paper, we investigate both of these strategies on deep models operating on point clouds. We propose \textit{PointMask}, a model-agnostic interpretable information-bottleneck approach for attribution in point cloud models. PointMask encourages exploring the majority of variation factors in the input space while gradually converging to a general solution. More specifically, PointMask introduces a regularization term that minimizes the mutual information between the input and the latent features used to masks out irrelevant variables. We show that coupling a PointMask layer with an arbitrary model can discern the points in the input space which contribute the most to the prediction score, thereby leading to interpretability. Through designed bias experiments, we also show that thanks to its gradual masking feature, our proposed method is effective in handling data bias.  

\end{abstract}

\section{Introduction}
%interpratibility methods in vision (2d images)
The performance of deep neural networks is usually measured based on their predictive behavior on a validation/test set. However, evaluating a model's performance on a single dataset can hardly capture its underlying behavior---even if the dataset is large enough~\cite{geirhos2020shortcut}. Therefore, having lucid explanations about predictions made by neural networks is critical for many applications. For example, in the image classification task, one cannot draw a reasonable conclusion solely based on predicted class 
probabilities ~\cite{geirhos2020shortcut} where variations in background or textures can completely change the predictions~\cite{beery2018recognition,rosenfeld2018elephant}. This critical flaw, which is mostly hidden, arises due to bias in training data~\cite{torralba2011unbiased}. Tasks such as image segmentation or object detection, on the other hand, where the output has a perceptible relation with the input, are inherently more explainable since one can infer if a model behaves anomalously by looking at a segmentation mask or a detected bounding box.

%possible applications of the above methods
Deep classifiers are effective in finding a few---but not most/all---discriminative variables in the input~\cite{geirhos2018imagenet}. Relying on a few and often correlated variables can lead to poor generalization when the variables are absent at test time due to a shift in data distribution~\cite{jo2017measuring,geirhos2020shortcut}. Uncovering biases in training data and inspecting whether a deep model converges to only a few input variables are two equally important advantages/applications of interpretability methods.

%move to point clouds
Despite the growing interest in interpreting 2D deep vision models, a scant effort has been made in interpreting deep networks processing point clouds, i.e., sparse order-invariant sets of interacting points representing 3D geometric data. Similar to images, point cloud datasets can be biased towards a specific pattern/feature. For example, a 3D sensor such as LiDAR may add a deliberate noise pattern to samples. In a multi-sensory setting, this can contribute to a \textit{fake} improvement of classification performance when all/most samples of a specific class are collected from a sensor with a particular and consistent noise pattern which is equivalent to ``context"~
\cite{beery2018recognition} or ``texture"~\cite{geirhos2018imagenet,baker2018deep} biases in images. An ideal model would learn multiple features/variables in a balanced way such that the absence of one feature at test time would not cause a drastic failure.

Interpretability methods can be categorized into three categories based on the training phase that they are applied to. Some methods are applied prior to training. Analyzing datasets for possible biases using clustering techniques are examples of such methods. Some other methods are applied during the training. Self-explanatory models which have interpretability modules are examples of these methods~\cite{zhmoginov2019information, zhang2018interpretable, fong2017interpretable, dabkowski2017real}. Finally, post-hoc techniques are applied to trained models~\cite{simonyan2013deep, selvaraju2017grad, smilkov2017smoothgrad, sundararajan2017axiomatic, springenberg2014striving, schulz2020restricting}. Our work falls into the second group.

In this paper, we focus on interpreting deep models that process point clouds with the ultimate goal of adding robustness against potential dataset biases. Our approach can be coupled with various network architectures without any constraints. We opt for simplicity and adopt the commonly used PointNet~\cite{qi2017pointnet} architecture for our study. Inspired by InfoMask~\cite{taghanaki2019infomask}, an information-bottleneck approach~\cite{alemi2016deep} for semi-supervised object localization, we design a model which detects/visualizes input variables that PointNet relies on the most to make predictions. We show that adding a masking layer to PointNet not only provides interpretability but also increases the model's robustness against \textit{bias} in training data and improves the model's performance on predicting the classes of randomly rotated objects. 
In summary, we make the following contributions:
\begin{itemize}
    \item We extend InfoMask~\cite{taghanaki2019infomask} to interpret deep models operating on point clouds. We call the introduced model \textit{PointMask} which learns to mask out input points with a negligible contribution to the model's predictions while encourages for exploring the majority of the input variables.
    \item We also introduce \textit{PointMap}, a variant of PointMask which instead of masking, learns to map the points into a new space, and study both models' effectiveness for unbiased point cloud processing.
    \item Finally, we show that removing extra input variables introduces a regularization effect which increases the model's robustness to random rotations.
\end{itemize}

\section{Related Work}
\textbf{Interpretability on point clouds.} Deep models introduced in the literature for processing point clouds primarily use visualizations to gain insights about the learned representations. In pioneer works such as PointNet~\cite{qi2017pointnet}, PointNet++~\cite{qi2017pointnet++},  and dynamic graph CNN~\cite{wang2019dynamic}, the intermediate representations are visualized by projecting them back into the 
point set space, whereas in \cite{zhao20193d, hassani2019unsupervised}, the evolution of learned representations is visualized through the training iterations. FoldingNet \cite{yang2018foldingnet} addresses interpretability from two aspects: (1) clear geometric interpretation by imposing a \emph{virtual force} to deform a 2D grid lattice onto a 3D object surface, and (2) visualizing gradual change of the folding forces.

To the best of our knowledge, C-PointNet~\cite{zhang2019explaining, huang2019claim} is the only relevant work focusing on the interpretability of point cloud models. It aggregates the point features learned by PointNet into a class-attentive global feature and generates class-attentive response maps to explain the decision making process in PointNet. Our approach, on the other hand, introduces a differentiable layer before the encoder that learns to mask out the points with negligible contributions by maximizing mutual information between the masked points and the class labels. It is noteworthy that our proposed module can be integrated into any other point set encoder without any constraints. We opt for simplicity and adopt the commonly used PointNet architecture in this work.

\textbf{Designing interpretable models.} Unlike the high interest in post-hoc interpretability methods~\cite{simonyan2013deep,selvaraju2017grad, smilkov2017smoothgrad, sundararajan2017axiomatic, springenberg2014striving, schulz2020restricting}, only a few works focus on adding interpretable components to deep models since it results in accuracy degradation on benchmark test sets. \cite{zhang2018interpretable} modified convolution layers using masks to obtain sharper feature maps. In \cite{zhmoginov2019information}, a hard attention mechanism based on mutual information is introduced that detects the most discriminative areas of the input. In the work done by \cite{fong2017interpretable}, a framework is proposed that learns masks to find parts of an image that influence the classifier's decision the most. Similarly, ~\cite{dabkowski2017real} apply a masking model to manipulate the scores of a pre-trained classifier by masking salient parts of the input. Nevertheless, these methods are all designed for RGB images whereas our method is specifically designed for point clouds. Moreover, the behavior of these models in the presence of data bias is unclear. We, on the other hand, analyze our method with different levels of input bias. Further, it is worth noting that our method does not lead to a decrease in overall classification accuracy.

% add a bit on interpretability methods which are related to causality (https://arxiv.org/pdf/1907.07165.pdf)

\section{Method}\label{sec:method}
\begin{figure}
     \centering
     \includegraphics[width=.55\columnwidth]{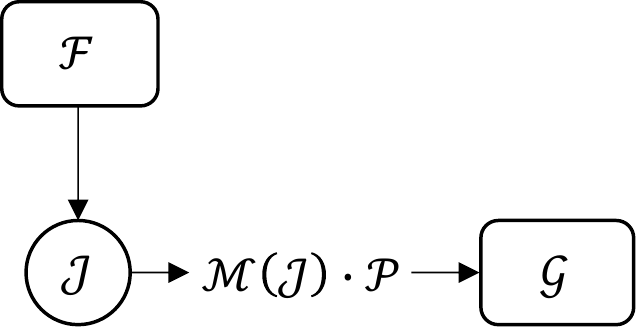}
     \caption{PointMask: the entire model is learned end-to-end.\\ $\mathcal{F} = p(\mathcal{J} | \mathcal{P} ; {\omega})$ and $\mathcal{G} = p(\mathcal{Y} | \mathcal{M}(\mathcal{J}).\mathcal{P} ; {\omega})$.}
     \label{fig:method}
\vskip -0.1in
 \end{figure}

Given a point set $\mathcal{P}=\left\{\mathcal{P}_{i} | i=1, \ldots, n\right\}\in \mathbb{R}^{n \times 3}$, where $\mathcal{P}_{i} \in \mathbb{R}^3$ denotes $(x, y, z)$ coordinates of point $i$, our goal is to calculate a set 
of variational importance probabilities $\mathcal{J}=\left\{\mathcal{J}_{i} | i=1, \ldots, n\right\} \in \mathbb{R}^n$ corresponding to $\mathcal{P}$ by optimizing:
\begin{align}
    \mathcal{L}(\omega)=I(\mathcal{J}, \mathcal{Y} ; \omega)-\alpha I(\mathcal{J}, \mathcal{P} ; \omega)
\end{align}
where $I$ is mutual information estimator, $\omega$ is network parameters, $\alpha$ is a scalar weight, and $\mathcal{Y}$ is class label of $\mathcal{P}$. We rewrite $I\left({\mathcal{J}}, \mathcal{Y} ; {\omega}\right)$ (and similarly$I(\mathcal{J}, \mathcal{P} ; \omega)$) as:
\begin{align}
I\left({\mathcal{J}}, \mathcal{Y} ; {\omega}\right) = 
\int p\left({\mathcal{J}}, \mathcal{Y} ; {\omega} \right)    \log \frac{p\left({\mathcal{J}}, \mathcal{Y} ; {\omega}\right)}{p\left({\mathcal{J}} ; {\omega}\right) p(\mathcal{Y} ; {\omega})} d\mathcal{P} d\mathcal{Y} 
\end{align}
We assume that the underlying distribution of $\mathcal{J}$ is a normal distribution $\mathcal{N}\left(\mu_{\mathcal{J}}, \sigma_{\mathcal{J}}\right)$ and learn the distribution parameters $\mu\in \mathbb{R}^{n}$ and
$\sigma \in \mathbb{R}^{n}$ using function $\mathcal{F}$ implemented as a feed-forward neural network. To sample $\mathcal{J}$, we use reparameterization trick, i.e., $\mathcal{J} = \mathcal{F} (\mathcal{P}, \epsilon) = \mu_{\mathcal{J}} + \sigma_{\mathcal{J}}\epsilon$. We define a masking function  $\mathcal{M}=\text{ReLU1}(\sigma(\mathcal{J})-t)$ to discard probabilities less than a threshold $t$ where $\sigma$ is sigmoid activation and ReLU1 is ReLU activation with upper-bound of 1. We multiply $\mathcal{M}$ by $\mathcal{P}$ to remove less critical points 
in the input and forward the remaining points to an point set classifier $\mathcal{G}$ which is PointNet in our case. The proposed model is shown in Figure~\ref{fig:method}.

The loss function consists of a classification loss and a regularization as follows:
\begin{align}\label{eq:loss}
    \mathcal{L}=\frac{1}{N} \sum_{n=1}^{N} \mathbb{E}_{\epsilon \sim p(\epsilon)}\left[-\log q\left(\mathcal{Y}_{n} | \mathcal{M}\left(\mathcal{P}_{n}, \epsilon\right)\right)\right]+  \nonumber  \\
    \alpha \mathrm{KL}\left(p\left(\mathcal{J} | \mathcal{P}_{n}\right) \parallel r(\mathcal{J})\right) \qquad \qquad 
\end{align} 
where $N$ is the number of training samples, $q(.)$ is the variational approximation function, $r(\mathcal{J})$ is the variational approximation to the marginal $p(\mathcal{J})=\int p(\mathcal{P}) p(\mathcal{J} | \mathcal{P}) d\mathcal{P}$, and KL(.$\parallel$.) is the Kullback–Leibler divergence. The scalar weight $\alpha$ controls the level of information that is passed through from the input. $\alpha$ is critical for encouraging the model to \textit{not} over-fit to a few input variables but to explore most/all of them leading to better generalization in a new environment where a few of the variables are absent.

Alternatively, we can change $\mathcal{M}$ from a masking function into a mapping function by modifying the dimension of $\mathcal{J}$ from $\mathbb{R}^n$ to $\mathbb{R}^{n\times 3}$, and treating it as a per-point translation function. Therefore, $\mathcal{M}=\text{ReLU1}(\sigma(\mathcal{J})-t)$ becomes $\mathcal{M}=\mathcal{J}$. This can be summed with $\mathcal{P}$ to map it into a new space as controlled by Equation~\ref{eq:loss}. We call this variant \textit{PointMap}.

\section{Experiments}

In the following subsections, we show PointMask's capability to interpret the classifier's decision by detecting the important points. We then examine how it behaves given a biased training set, and finally, we test whether the masking step of PointMask helps to differentiate similar objects which are often misclassified.

We evaluate our model on ModelNet40 and ModelNet10 shape classification benchmarks~\cite{wu20153d}. ModelNet40 consists of 12,311 CAD models from 40 man-made object categories, split into train and test sets of sizes 9,843 and 2,468, respectively. ModelNet10 is a subset of ModelNet40 and contains 4899 samples split into 3991 training and 908 test samples. 

\subsection{Interpreting by Detecting Important Input Variables}

\begin{figure}
     \centering
     \includegraphics[width=0.99\columnwidth]{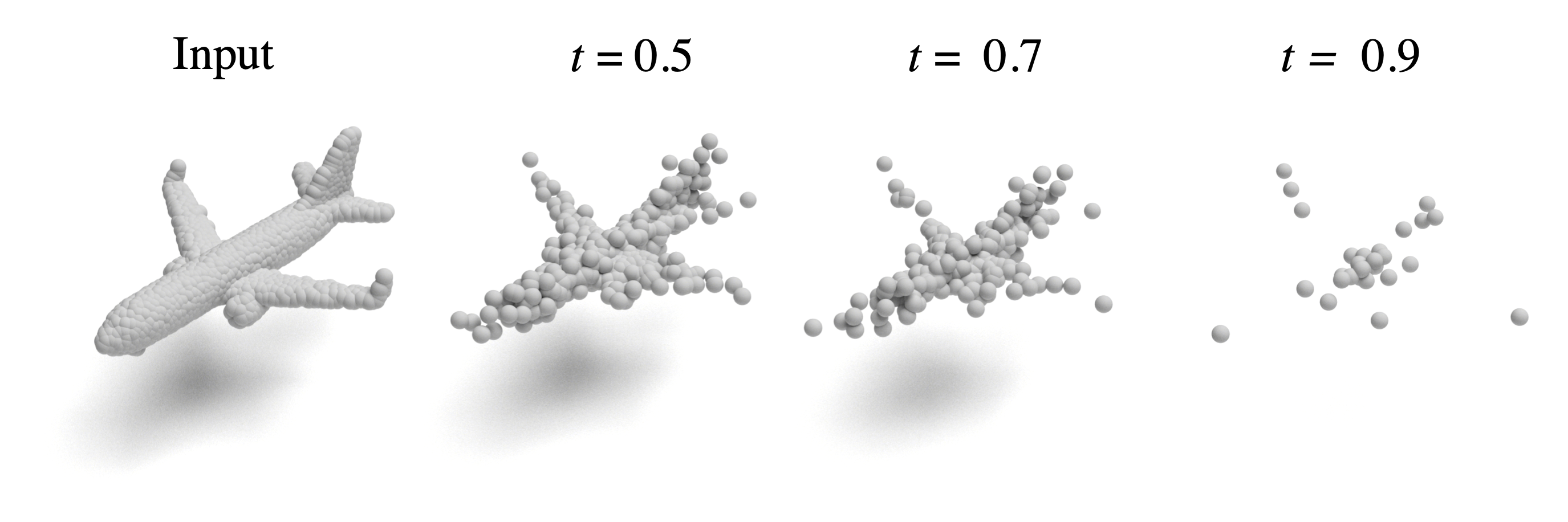}
     \caption{PointMask with different threshold values; increasing the threshold leads to detecting key elements without using any prior knowledge.}
     \label{fig:threshold}
\vskip -0.2in
 \end{figure}

The first goal of this work is to visualize/detect important input variables to answer the question: \textit{``Which subset(s) of a given input point cloud contribute(s) most to the final prediction?''} To this end, we propose to place a trainable variational masking layer before PointNet which learns to remove less critical points (see section ~\ref{sec:method}). As shown in Figure~\ref{fig:threshold}, increasing the masking threshold improves the detection of the points that correlate the most with the object class. 

We also compare PointMask with PointNet trained with different types of data augmentation and tested on aligned samples. Results reported in Table~\ref{tab:aug} suggest that PointMask in addition to providing interpretability, outperforms PointNet on the ModelNet40 dataset. We observed that augmenting the training set with random rotations results in an accuracy drop of $\sim 1.0\%$ on both 
models lose (see section~\ref{sec:rot}).

\begin{table} \label{tab:aug}
\setlength{\tabcolsep}{4pt}
\caption{ModelNet40 classification results for different models trained with data augmentation and tested on aligned samples. J, R1, R3, and A refer to jitter, rotation along a single axis, rotation along all 3 axes, and aligned samples respectively.}
\centering
\begin{tabular}{lccc}
\toprule
         & \multicolumn{3}{c}{Trained with - Tested on} \\ \hline
         & J - A         & JR1 - A       & JR3 - A      \\ \hline
PointNet & 89.61       & 89.45       & 88.39      \\
PointMask & \highlight{89.73}       & 89.37       & \highlight{88.88}       \\
\bottomrule
\end{tabular}
\end{table}

\subsection{Unbiased Feature Learning}
\begin{figure}%[htbp]
     \centering
     \includegraphics[width=.7\columnwidth]{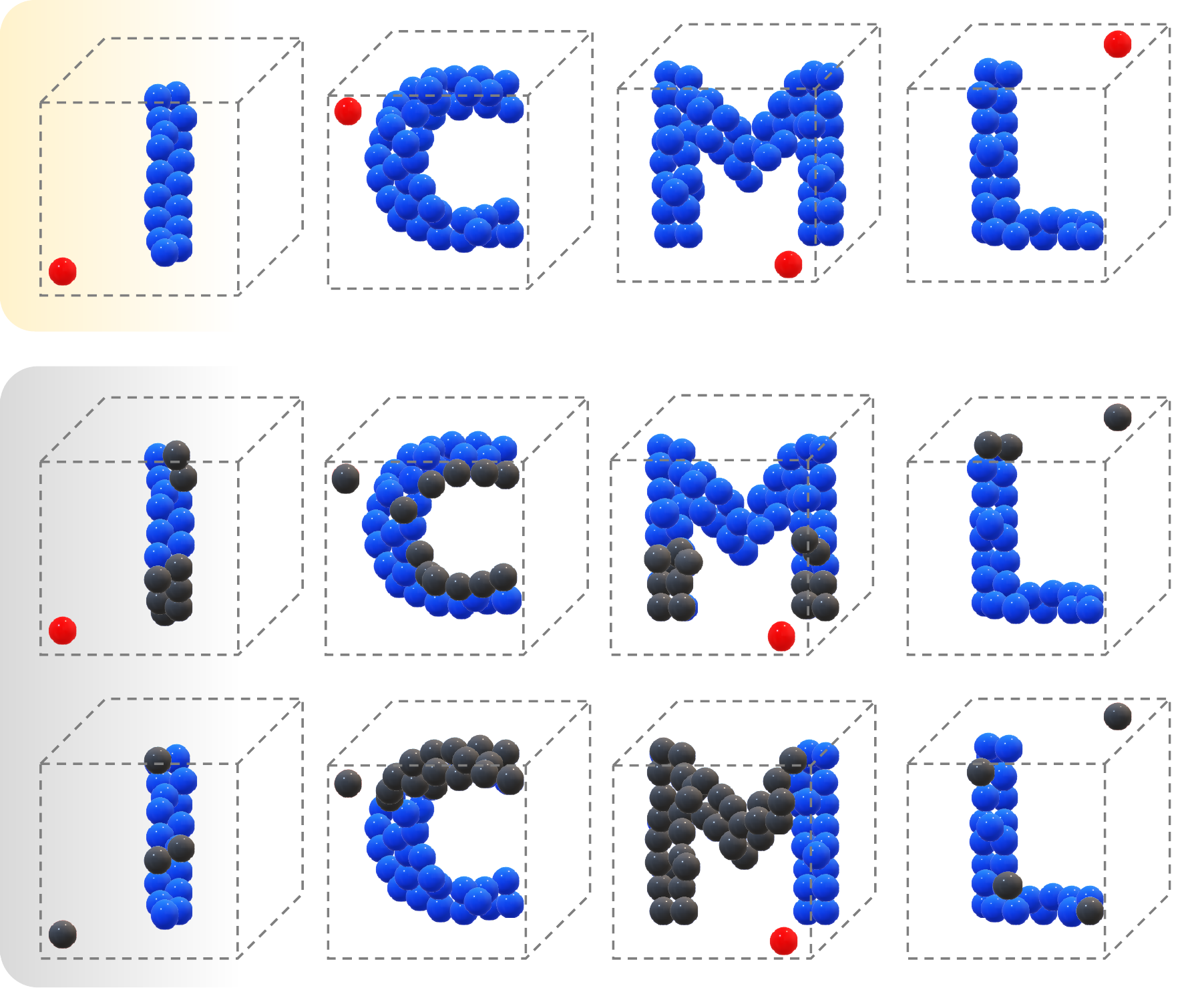}
     \caption{A toy example depicting single point bias. The blue, black, and red points represent input, masked, and bias points, respectively. The first row shows samples with bias from different categories while the second and third row visualize the same samples which are randomly masked. In some samples bias and some parts of the objects have been masked together.}
     \label{fig:icml}
 \vskip -0.2in
 \end{figure}

\begin{figure}[htbp]
     \centering
     \includegraphics[width=.6\columnwidth]{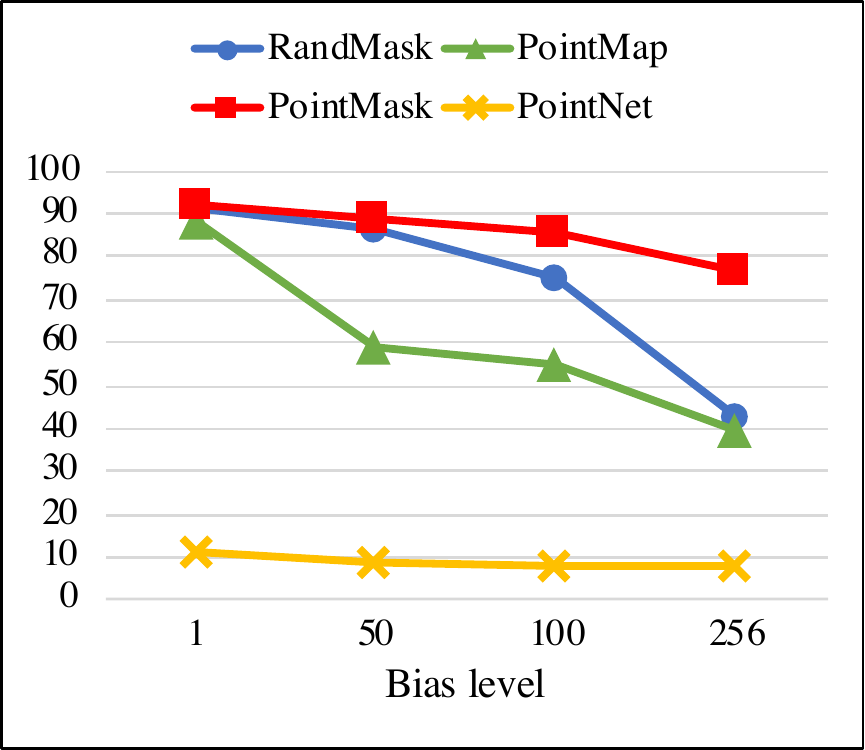}
     \caption{ModelNet10 classification results with different levels of bias added during training and removed at test time.}
     \label{fig:masking_results}
     \vskip -0.2in
 \end{figure}

\begin{figure*}
     \centering
     \includegraphics[width=0.86\textwidth]{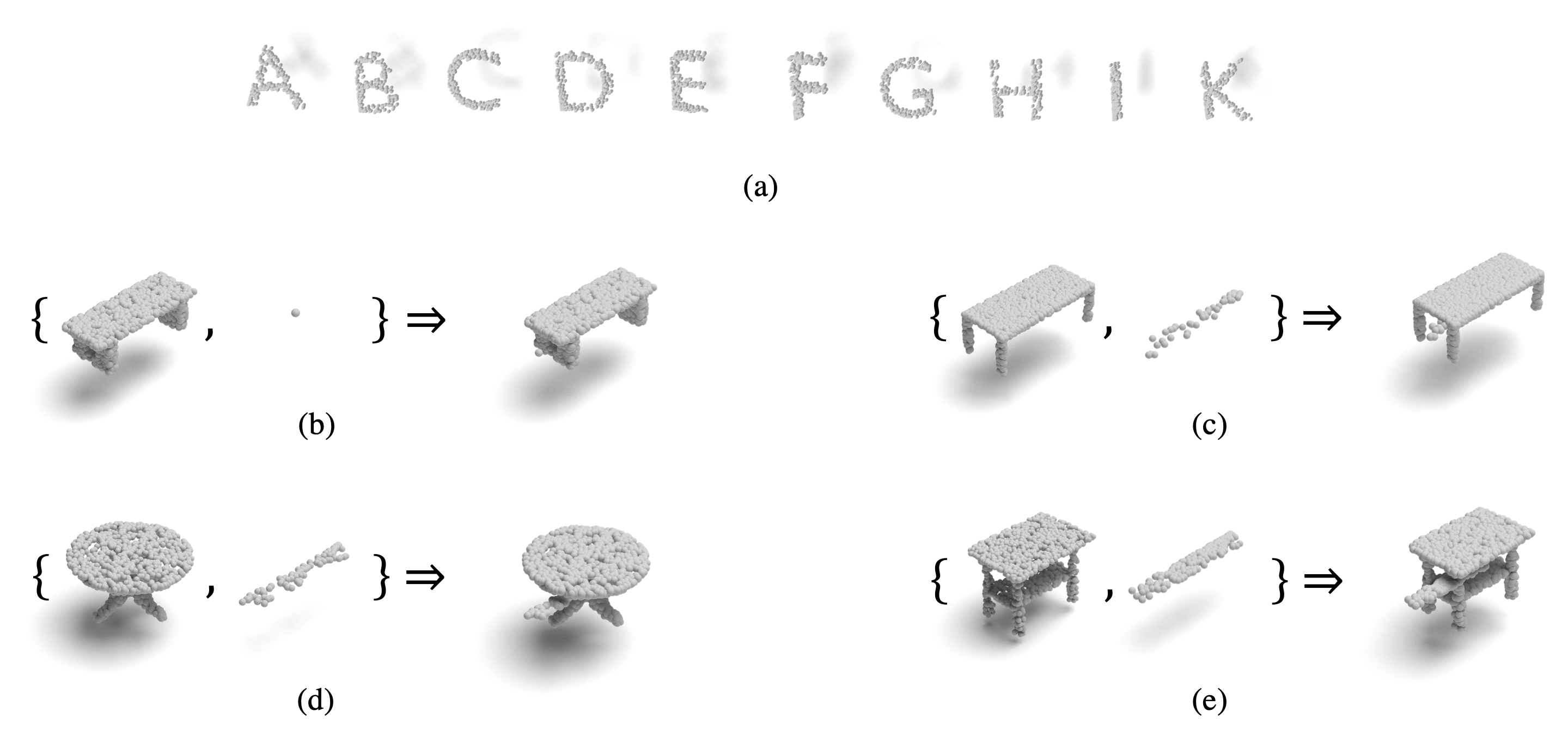}
     \caption{Generated biased samples from ModelNet10 dataset. (a) point cloud alphabet set which we add to objects as bias, (b)--(e) biased samples with 1, 50, 100, and 256 bias points, respectively.}
     \label{fig:noise_samples}
     \vskip -0.2in
 \end{figure*}

A deep classifier might converge to one or a few input variables/patterns such as texture, shape, etc. To study this on point clouds, we design a controlled experiment and systematically inject unique and constant patterns to each object category. The red points placed in different coordinates in the first row of Figure~\ref{fig:icml} show the systematic patterns that are added to each class during the training. We perform a similar experiment using ModelNet10 dataset~\cite{wu20153d}. We add a point cloud alphabet (Figure~\ref{fig:noise_samples}, (a)) to each category of shapes, e.g., letter ``I'' is concatenated with all desks where ``I'' is sampled with different number of points $\{1, 50, 100, 256\}$ to increase the strength of the pattern (Figure~\ref{fig:noise_samples} (b-e)). The goal here is to test whether PointNet attends to any part of the object or just overfits to the unique pattern. With the presence of patterns at test time, PointNet obtains $\sim 100\%$ classification accuracy. However, as shown in Figure~\ref{fig:masking_results}, when the patterns are removed at test time, its performance drops to random accuracy, even with a single point pattern. This indicates that regardless of the other 2048 points, PointNet only relies on a single point bias pattern. 

As a simple remedy for constant bias patterns, we randomly mask n\% of the points of each object where $n\in[10,70]$. The hope here is that some/all bias patterns might be removed in some training samples (see the second and third rows of Figure~\ref{fig:icml}), thereby forcing a model to look at other variables, i.e., object points. As shown in Figure~\ref{fig:masking_results}, this approach turns out to be effective for weak bias patterns, however, when the bias level increases to 256 points, its classification accuracy on ModelNet10 decreases to 42\%. 

Next, we train PointMask and PointMap using the same biased train set and evaluate them with a bias-free test set. As shown in Figure~\ref{fig:masking_results}, PointMask performs reasonably well despite the increase in the bias level, e.g., for the strongest bias level, it achieves an accuracy of 76\% which is and 34\% absolute improvement over RandMask. The intuition for this improvement is that PointMask samples masks from a normal distribution and the KL divergence term in Equation~\ref{eq:loss} enforces the masks to have a minimum amount of information about the input points; on the other hand, the classification loss enforces the sampled masks to have maximum information about class labels. This min-max game prevents the model from over-fitting to a premature solution. Therefore the model has the chance to explore a diverse set of masks per object and factors of variation during its convergence process. 

\begin{figure}
     \centering
     \includegraphics[width=0.99\columnwidth]{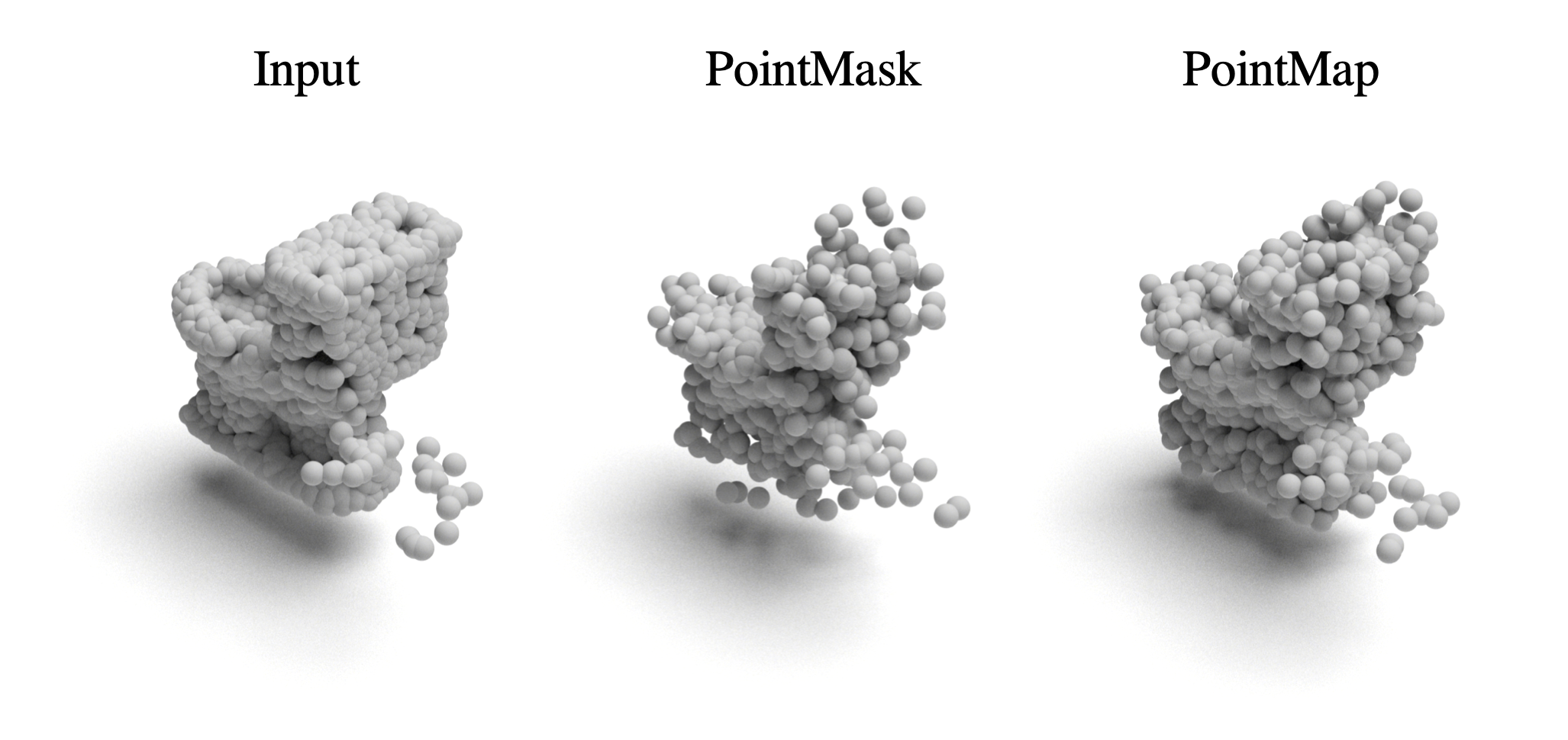}
     \caption{Visualization of processed biased samples from ModelNet10 with different methods.}
     \label{fig:bias_remove}
     \vskip -0.2in
 \end{figure}

We observed that for strong bias patterns, PointMap is not as effective as PointMask. This is because, in order to obliterate the bias, PointMap has to apply a large translation per point and since there is no prior knowledge on the ``objectness" of the samples, this translations will morph the object to something else. Therefore, there will be no unique features across the classes left for the model to use to improve classification accuracy. This phenomenon is visualized in Figure~\ref{fig:bias_remove} where PointMask is able to eliminate the bias by removing points, whereas for the object processed by PointMap, the bias is still present/detectable.

\subsection{Regularization Effect on Rotated Objects}\label{sec:rot}
We also investigated whether translating/masking irrelevant points can lead to a better classification accuracy once models are tested on rotated objects. We noticed when PointNet is trained and tested with randomly rotated samples along all three $\{x, y, z\}$ axes, its accuracy drops from 89.6\% to 76.9\% even though it is augmented with random rotations during training. On the other hand, PointMask improves performance from 76.9\% to 82.2\%. We speculate this is partly because of improper regularization in PointNet, i.e., when we removed dropout layers from PointNet, the accuracy increased from 76.9\% to 81.1\% (see Table~\ref{tab:r3}). Nevertheless, even with this modification, it is still outperformed by PointMask.  

\begin{figure}
     \centering
     \includegraphics[width=.7\columnwidth]{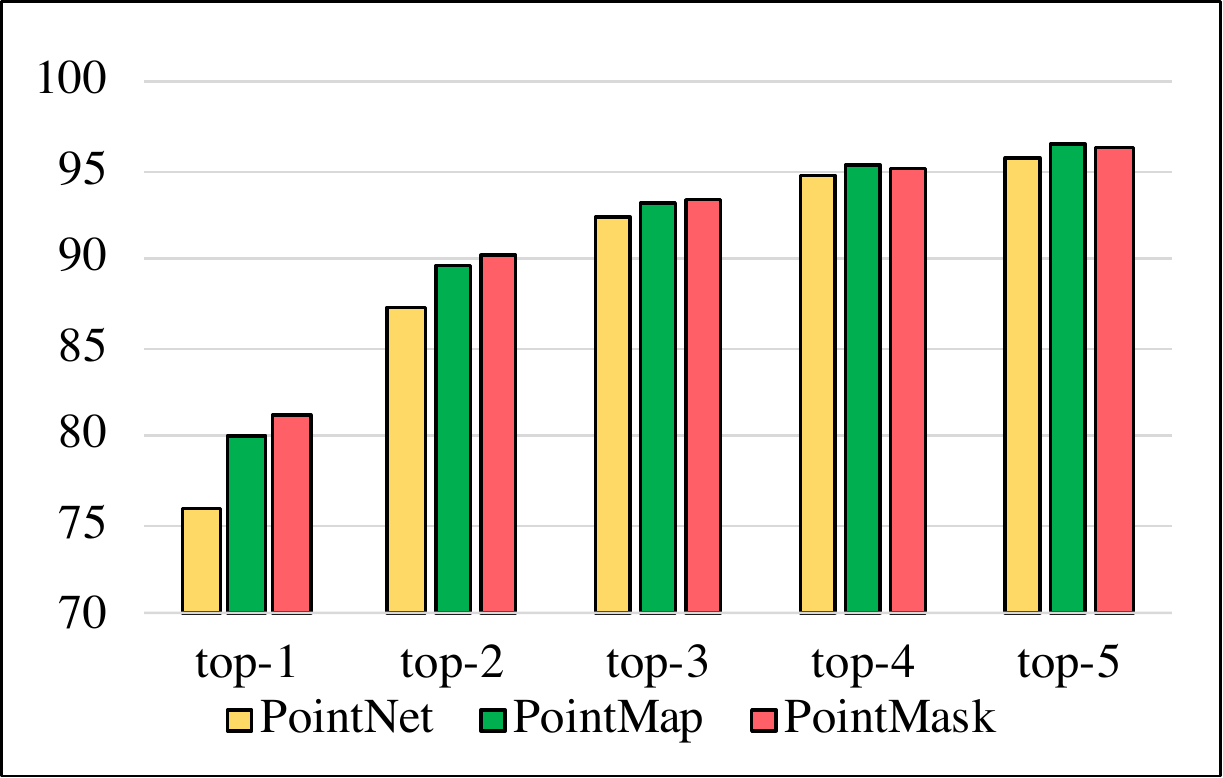}
     \caption{Top-n classification accuracy on ModelNet40.}
     \label{fig:barchart}
 \end{figure}

In addition, input rotations can confuse the model. Some object categories in ModelNet40 (e.g., \texttt{bookshelf} and \texttt{bed}) look considerably similar when viewed from similar orientations. We speculate that PointMap and PointMask help to differentiate similar objects by making them look different via point translation and removal, respectively, even if they are in the same pose. In Figure~\ref{fig:barchart}, we show that by increasing $n$ in top-n accuracy, all models achieve almost similar performance. However, looking at top-1 and top-2 accuracy indicates that PointNet was confused among the top few classes, whereas PointMap and PointMask obtained reasonable performance.

 hypothesize
We further examine the per-class accuracy by different models using a confusion matrix. As shown in Table~\ref{tab:classconfusion}, PointMap and PointMask improve accuracy on several classes. We observe that there is a big confusion between \texttt{flowerpot} (0.0) and \texttt{plant} (50.0) classes in PointNet, which goes down to 40.0 in both PointMap and PointMask. Similarly, the confusion of the \texttt{cup} (35.0) class is highest with the \texttt{vase} (45.0) class, and goes down to 25.0 and 30.0 in PointMap and PointMask, respectively. We observe similar trends in other poorly performing classes as well.
\begin{table}\label{tab:classconfusion}
\caption{Classification accuracy for lowest scoring ModelNet40 classes.}
\centering
\begin{tabular}{lccc}
\toprule
Class               & PointNet & PointMap           & PointMask\\
\midrule
\texttt{flowerpot}  & 00.0    & 20.0               & \highlight{25.0} \\
\texttt{cup}        & 35.0    & 50.0               & \highlight{55.0} \\
\texttt{radio}      & 30.0    & 40.0               & \highlight{60.0} \\
\texttt{wardrobe}   & 5.00    & \highlight{0.40}   & \highlight{40.0} \\
\bottomrule
\end{tabular}
\end{table}

\begin{table}\label{tab:r3}
\caption{ModelNet40 classification results for different models trained and tested with randomly rotated objects. Rotation here can be about any random 3D axis.}
\centering
\begin{tabular}{lcc}
\toprule
         & Original & Without dropout   \\ \hline
PointNet & 76.87  & 81.13 \\
RandMask & 75.20  & 80.64 \\
PointMap  & 81.05  & NA      \\
PointMask & \highlight{82.18}  & NA      \\ 
\bottomrule
\end{tabular}
\end{table}

\begin{figure}[]
     \centering
     \includegraphics[width=0.99\columnwidth]{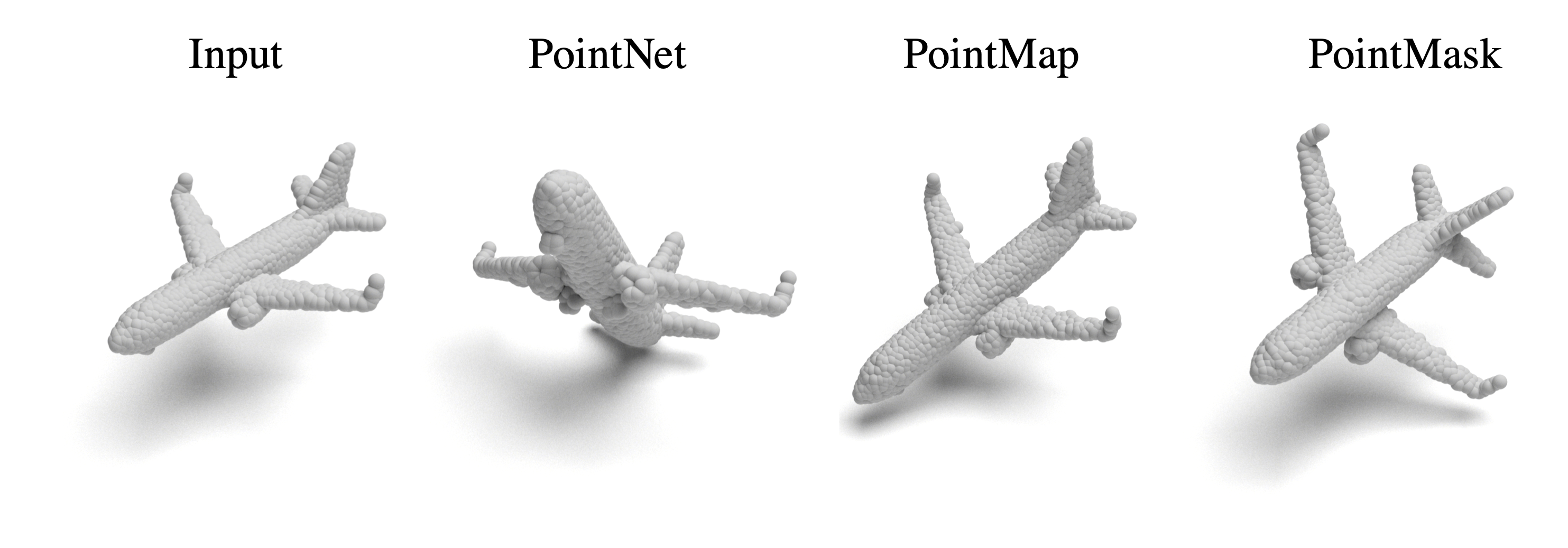}
     \caption{The output of input transformation module (T-Net and rotation) in PointNet for different methods.}
     \label{fig:roation}
 \end{figure}

In Figure~\ref{fig:roation}, we visualize the output of the input transformation module in PointNet which consists of the T-Net and rotation operation. As shown, different rotations are obtained when the same input is processed by different methods.

\subsection{Implementation Details}
For all models, we set the number of points to 2048 and their dimension to 3, i.e., Euclidean coordinates. We optimized the models for 500 epochs using Adam~\cite{kingma2014:adam} with learning rate set to 0.0001, and mini-batch size of 32. We provide pseudocode for PointMask in Listing~\ref{code:pointmask}.

\begin{lstlisting}[language=Python,caption=Pseudocode for PointMask.,label=code:pointmask]
def mask_relu(x, threshold = 0.5):
    x = sigmoid(x)
    x = relu(x - threshold)
    x = clamp(x, min=0, max=1)
    return x

def kl_divergence(mu, log_var):
    tmp = 1.0 + log_var - mu * mu - exp(log_var)
    kl_batch = -0.5 * sum(tmp, axis=-1)
    return alpha * mean(kl_batch)

# PointMask
num_points = 2048 
feature_dim = 3  # (x, y, z) coordinates
@($\mathcal{P}$)@ = ... # pointcloud of shape (num_points, feature_dim)
l = Conv1DReLUBN(channels=64)(@($\mathcal{P}$)@)
l = Conv1DReLUBN(channels=128)(l)
l = Conv1DReLUBN(channels=1024)(l)
l = MaxPool1D(pool_size=2048)(l)

# Reparameterization trick
mu = Conv1D(channels=2048)(l)
log_var = Conv1D(channels=2048)(l)
loss_kl = kl_divergence(mu, log_var)
sigma = exp(0.5 * log_var)
eps = random_normal(stddev=1.0, shape=(num_points, 1, 2048)))
eps = sigma * eps
@($\mathcal{J}$)@ = mu + eps

# Masking input points
@($\mathcal{M}$)@ = mask_relu(@($\mathcal{J}$)@)
masked_input = @($\mathcal{P}$)@ * @($\mathcal{M}$)@ 

# Forward masked points to PointNet
pred = @($\mathcal{G}$)@ (masked_input)  # G = PointNet
loss_ce = cross_entropy(pred, ground_truth)
@($\mathcal{L}$)@ = loss_kl + loss_ce 

\end{lstlisting}

\section{Conclusion}
We introduced PointMask, an information-bottleneck interpretability layer that can be integrated into any point cloud classification model. We showed that PointMask not only provides interpretability but also brings robustness against deliberate bias patterns which might not be perceptible to humans. We also demonstrated that the regularization effect of the proposed model is more effective compared to dropout. Finally, we showed that our method reduces confusion among classes that might be similar when arbitrarily rotated. As a future direction, we are planning to apply the proposed method to key point/landmark detection in point clouds.

\section*{Acknowledgements}
We would like to thank Aditya Sanghi and Ara Danielyan for their useful input through the course of this research.


\bibliography{example_paper}
\bibliographystyle{icml2020}

\end{document}